\documentclass[conference]{IEEEtran}
\IEEEoverridecommandlockouts
\usepackage{cite}
\usepackage{amsmath,amssymb,amsfonts}
\usepackage{algorithmic}
\usepackage{graphicx}
\usepackage{textcomp}
\usepackage{xcolor}
\usepackage{subfigure}
\usepackage{graphicx}
\usepackage{float} 

\def\BibTeX{{\rm B\kern-.05em{\sc i\kern-.025em b}\kern-.08em
    T\kern-.1667em\lower.7ex\hbox{E}\kern-.125emX}}
\begin{document}

\title{Real-Time Elderly Monitoring for Senior Safety by Lightweight Human Action Recognition}

\author{\IEEEauthorblockN{Han Sun, Yu Chen}

\IEEEauthorblockA{Department of Electrical and Computer Engineering, Binghamton University, Binghamton, NY 13902, USA\\
\{hsun28, ychen\}@binghamton.edu}
}

\maketitle

\begin{abstract}
With an increasing number of elders living alone, care-giving from a distance becomes a compelling need, particularly for safety. Real-time monitoring and action recognition are essential to raise an alert timely when abnormal behaviors or unusual activities occur. While wearable sensors are widely recognized as a promising solution, highly depending on user's ability and willingness makes them inefficient. In contrast, video streams collected through non-contact optical cameras provide richer information and release the burden on elders. In this paper, leveraging the Independently-Recurrent neural Network (IndRNN) we propose a novel Real-time Elderly Monitoring for senior Safety (REMS) based on lightweight human action recognition (HAR) technology. Using captured skeleton images, the REMS scheme is able to recognize abnormal behaviors or actions and preserve the user's privacy. To achieve a high accuracy, the HAR module is trained and fine-tuned using multiple databases. An extensive experimental study verified that REMS system preforms action recognition accurately and timely. REMS meets the design goals as a privacy-preserving elderly safety monitoring system and possesses the potential to be adopted in various smart monitoring systems. 
\end{abstract}


\section{Introduction}
According to the U.S. Department of Health and Human Services (HHS), the population age 65 and over has increased from 37.2 million in 2006 to 49.2 million in 2016, and is is expected to nearly double to 98 million in 2060 \cite{AoA2017Profile}. Of these, about 28\% (13.8 million) of non-institutional seniors live alone. With the unprecedented increasing of population aging and more elders living alone, care-giving from a distance becomes a compelling need, particularly for safety. A major threat to elders living alone is health problems such as falls and unconscious. 
Other chronic diseases, such as hypertension and hypoglycemia, also have certain behavioral manifestations which usually manifest as headache, chest pain, staggering, vomiting, etc \cite{erickson2004relationship}. Timely detection and alarming of these abnormal situations are essential for their safety and health. 


Nowadays, there are many technologies applicable for remote safety monitoring and action recognition in healthcare services. Autonomous wearable sensors are often considered to detect dangerous actions like suddenly falls \cite{mubashir2013survey}. However, wearable devices are inconvenient to put on and take off, and users often forget to wear them \cite{foroughi2008intelligent}. Limited battery life is also put extra burden to users. Actually the effectiveness of wearable sensors is highly depending on the capability and willingness of the user. Meanwhile, video streams collected through non-contact optical cameras provide a richer information and release the burdens on elders. But the convenience is accompanied by the risk of privacy violations. 

Human skeleton is widely used in action recognition \cite{ke2017new}. In contrast to raw video streams, skeleton images are privacy-preserving by nature since a human body is represented as a coordinate of joints and background distractions are removed. Even if a skeleton image is leaked, it does not spill much personal information into the wild cyber-space.

Skeleton-based human action recognition (HAR) is generally considered as a time series processing problem \cite{li2014prediction}. Human actions in a period of time can be represented as a matrix with high auto-correlation inside. Therefore, deep learning (DL) methods that take advantage of the auto-correlation have achieved good results. The skeleton joints can be encoded into multiple 2D pseudo-images and then feed to Conventional Neural Networks (CNNs) to learn useful features \cite{xu2018ensemble, yang2019make}. But CNN itself is suffering from the problem of size and speed.
Graph Conventional Networks (GCN) \cite{yan2018spatial} firstly built a saptio-temporal graph, in which the joints are represented as graph vertices and the natural connectivity of human body structure and time is mapped to the graph edges. However, GCN-based HAR systems face the problem of data transforming. Time Pyramid performs well in analysing three-dimension (3D) geometric relationships but it is generally restricted by the width of the time windows \cite{luo2013group}. 

A novel two-stream recurrent neural network (RNN) is proposed to model both temporal dynamics and spatial configurations for skeleton data \cite{wang2017modeling}. RNNs have advantages in time-consuming processes when dealing with long sequences. But due to the recurrent connections with repeat multiplication of the recurrent weight matrix, the training of RNNs suffers from the gradient vanishing and exploding problem. Recently, Independently-Recurrent Neural Network (IndRNN) has been proposed to solve the problem of exploding gradient retains the advantages of RNN in processing auto-correlated data\cite{ge2019human}. In addition, IndRNN is able to achieve approximate recognition result with more concise structure compared with others\cite{li2018independently}.


In this paper, a novel Real-time Elderly Monitoring for senior Safety (REMS) scheme is proposed leveraging a real-time skeleton image action recognition system based on IndRNNs. The major contribution of this work includes the following:
\begin{itemize}
    \item An IndRNN based action recognition system is proposed that is able to sound alarm timely when emergency action occurs.
    \item Continuous real-time monitoring is enabled using a sliding window method that divides the input video stream into a sequence of overlapped short video segments to reduce the processing time for each segment.
    \item REMS achieved a better performance with a lightweight design, which makes REMS affordable for household use with very constrained computational resources.
\end{itemize}

The rest of this paper is organized as follows. Section \ref{background} provides necessary background knowledge and gives a brief overview of related work. Section \ref{system} introduces the methodology and the system framework of REMS. Section \ref{result} reports experimental results and a comparison study with several state-of-the-art methods. Section \ref{Conclusion} concludes this paper with some discussions on the on-going efforts, specifically in terms of security and privacy.

\section{Related Work}
\label{background}

\subsection{Action Recognition in Healthcare}
According to the technical approaches, recent human activity monitoring solutions roughly belong to two categories, wearable device-based and image processing-based \cite{yadav2021review}.

Wearable devices based HAR systems use variant types of sensors to collect human data for analysis, such as inertial sensors (e.g. accelerometers, gyroscopes, magnetometers), GPS, heart rate sensors, etc. \cite{bianchi2019iot}. For instance, five dual-axis accelerometers are installed on arms, hips, knees, and ankles to identify daily behaviors including walking, jumping, an overall accuracy rate of 84\% is achieved using a decision tree algorithm \cite{bao2004activity}. 
Falling can be detected in real-time by collected fall and non-fall data-sets \cite{aziz2017validation}. Five younger and 19 elder persons went on their everyday work by wearing accelerometers. Ten unexpected falling is collected during total 500 hours of data acquiring. These HAR methods using single or multiple acceleration sensors have achieved encouraging results. However, the recognition accuracy is highly correlated with the sensor positions, and most of the recognition is performed offline, which does not meet the real-time requirements. In addition, as only few recognition actions can be discriminated in real-time, their utility is limited.

With the widespread application of mobile robots in indoor scenes, ultra-wideband impulse-radar technology has also been applied to indoor positioning and motion capture, especially to capture dynamics in indoor mobile traffic scenes, such as hospitals and IoT factories. However, indoor UWB positioning technology can lead to inaccurate positioning due to measurement errors caused by obstacles and non-line-of-sight (NLOS)\cite{ma2022improved}.

Besides the applications in healthcare area, video surveillance has been widely adopted to serve many different purposes such as public safety \cite{xu2020blendsps}, defense \cite{chen2016real}, and smart transportation \cite{chen2016dynamic}. HAR based on visual processing normally uses image devices installed on predetermined points of interest. For example, a camera installed in the living room where an elder spends most of the time each day and a computer for intelligently analysis on the captured human actions in video or image sequences. 
By replacing the dense sampling with random sampling, the number of sampled patches to be processed is reduced such that the efficiency is increased to recognize human action in real-time \cite{shi2013sampling}. 

Although there are numbers of research that focuses on the RGB images acquired though cameras and achieves good recognition results, there are still shortcomings. The cameras must be installed in a specific environment, hence the viewing edge is usually limited. Also, image acquisition occupies a large amount of memory, therefore the hardware device is required to have large storage space and sufficient processing capabilities. The most important issue is, the data collected by cameras contains a lot of personal information while most computing intensive image/video processing tasks are outsourced to remote servers. The high risk of personal information leakage in the transmission of the raw images poses a non-negligible threat to privacy preservation \cite{fitwi2021privacy}. 

\subsection{Skeleton Features Extraction}
With the development of 3D depth camera, the 3D node positions of a human skeleton can be quickly obtained from the depth image. At each moment, the skeleton corresponds to the positions of the joint points of a human body. Recent years have witnessed a fast development of HAR techniques based on depth information in multiple fields including smart home, games, and human-computer interaction \cite{jalal2012depth}. Compared with RGB images, skeleton image is less susceptible to appearance factors.

The human skeleton is defined as a schematic model describing torso, head and limbs. The parameters and motion information can be used to represent gesture and pose. In 3D skeleton-based action representation, an action is a collection of 3D position-time series. However, differences in reference coordinate systems and recording environments, plus the differences among variant human motion styles, lead to errors if the HAR processing depends only on joint coordinate representing.
Therefore, many different implementations on representation method of skeleton matrix are introduced, such as the pose of human body is normalized by length of shoulders and torso \cite{wei2013concurrent}, the skeleton data is centered on the hip joint to establish coordinates \cite{vemulapalli2014human} such that the data can be rewritten according to a new coordinate system.

\section{REMS Methodology and System Architecture}
\label{system}

\begin{figure*}[ht]
  \centering
  \includegraphics[width=0.9\linewidth]{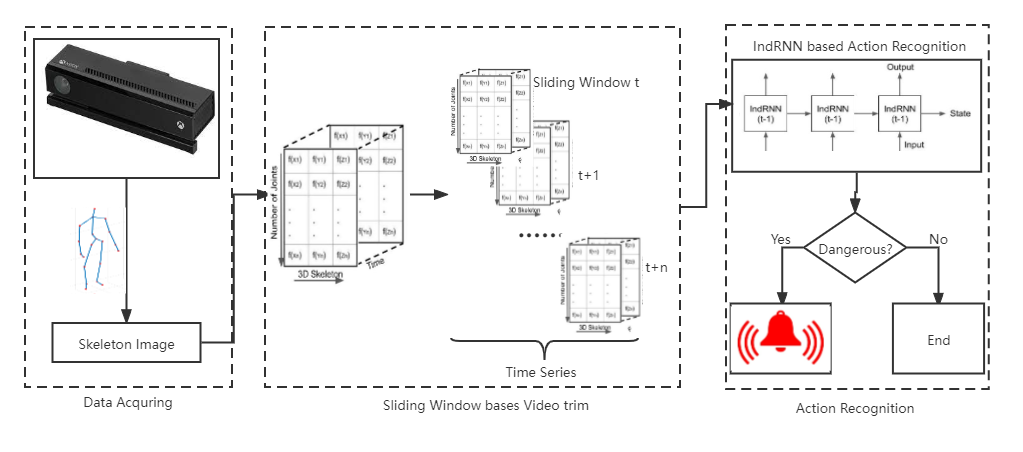}
  \caption{An overview of the REMS scheme.}
  \label{fig:REMS}
\end{figure*}

Figure \ref{fig:REMS} shows the complete algorithm flow of REMS system. The Kinect V2 camera is adopted to establish the human body 3D-skeleton image. The untrimmed 3D skeleton videos are cut into a overlapped continuous video stream using a sliding window method. The processed 3D skeleton video stream will fed to the IndRNN based action recognition system, which will analyze the video, identify suspicious activities, and sound an alarm when an emergency situation is detected.

\begin{figure}[ht]
\centering
\label{ske}
  \includegraphics[width=0.65\linewidth]{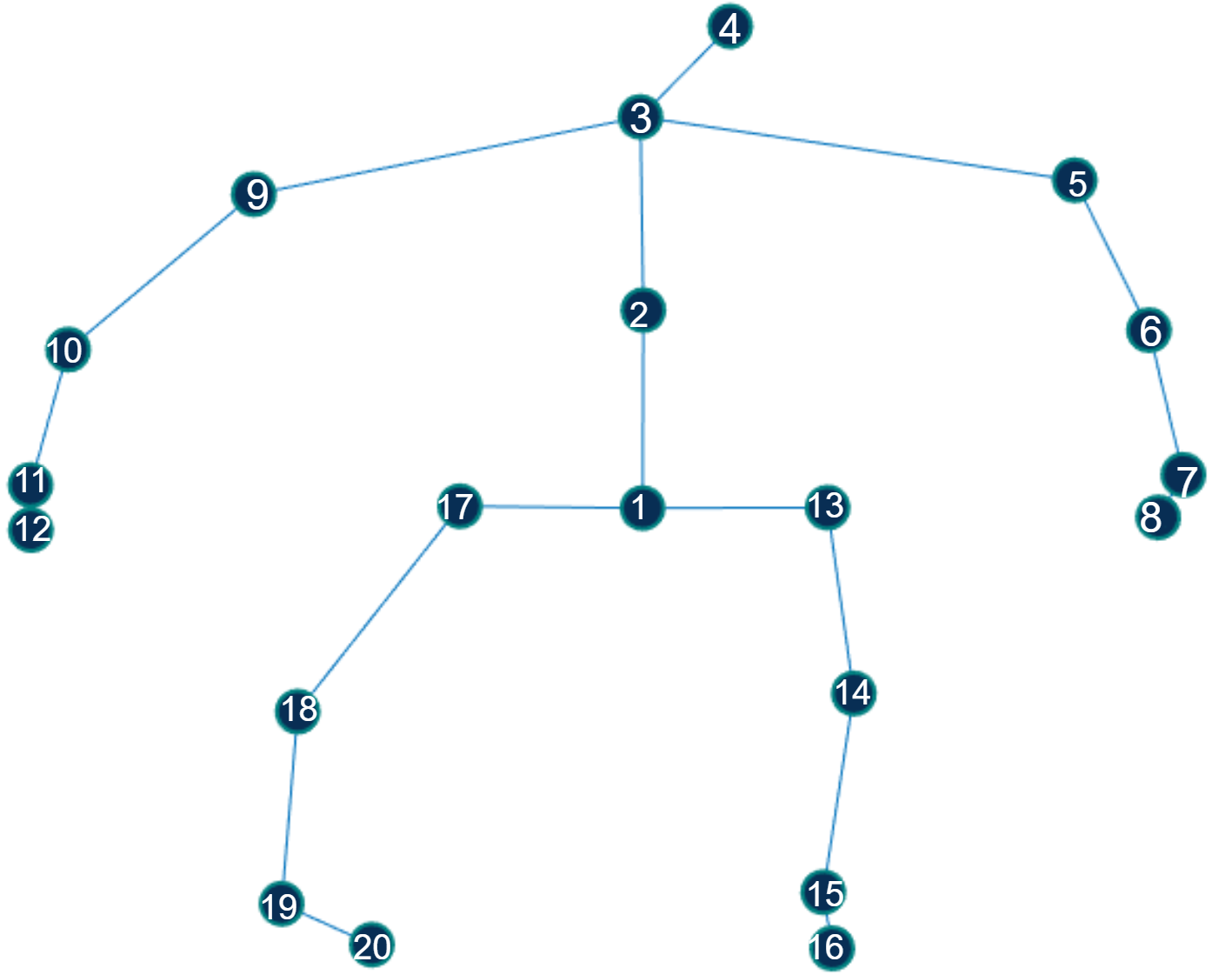}
  \caption{Configuration of the 20 joint in skeleton image, The labels of the joints are: 1.hip center, 2.middle-spine, 3.shoulder center, 4.Head, 5.Left-shoulder, 6.Left-elbow, 7.Left-wrist, 8.Left-hand, 9.Right-shoulder, 10.Right-elbow, 11.Right-wrist, 12.Right-hand, 13.Left-hip, 14.Left-knee, 15.Left-ankle, 16.Left-root, 17.Right-hip, 18.Right-knee, 19.Right-ankle, 20.Right-foot.}
  \label{fig:meth}
\end{figure}

\subsection{3D Skeleton Image Feature Extraction}
Kinect sensor is able to construct simplified human skeleton model by using 20 key points information instead of using total 206 bones. Figure \ref{fig:meth} presents the 20 key skeletal joints that sufficiently describe the process of general human movement. In this model, the spatial coordination information of each joint point is set as $P(x,y,z)$, where $x$ and $y$ are the abscissa and ordinate respectively, and the $z$ dimension represents the distance from the human body to the camera. During the movement, the relative positions among the joints change. For example, in the waving action, the joint of wrist is initially under the joint of shoulder. But as the movement progresses, the wrist joint moves above the shoulder joint. At the same time, the joints represent torso are always relatively stable. Therefore, to better represent the offset of the joint points of the limbs relative to the hip and remove the effect of distance from the camera, taking the central node of the hip as the central origin, the calculation formula for obtaining the initial spatial position feature is as follows:

\begin{equation}
    f = p_{n}-p_{hip}(n = 2,3,....N).
\end{equation}
where $p_{n}$ represents other nodes except the hip joint, and $p_{hip}$ is the hip-center joint. The representation in a 3D space is:
\begin{equation}
    \left\{\begin{matrix}
\Delta x_{n}^{m} = x_{n}^{m} - x_{h}^{m}\\ 
\Delta y_{n}^{m} = y_{n}^{m} - y_{h}^{m}\\ 
\Delta z_{n}^{m} = z_{n}^{m} - z_{h}^{m}
\end{matrix}\right. 
\end{equation}
\begin{equation}
\label{e3}
f_{x}^{m} = [\Delta x_{1}^{m}, \Delta x_{2}^{m}, .....\Delta x_{n}^{m}]
\end{equation}
The difference between the $X$ coordinates of the remaining 19 points in the $m$\_th frame and the center point can be obtained by Eq. \ref{e3}. In the same way, the difference in $Y$ axis $f_{y}^{m}$ and the difference in $Z$ axis $f_{z}^{m}$ can be obtained. The three dimensional coordinates can be connected to the feature vector of the current frame: $f_{m} = [f_{x}^{m},f_{y}^{m},f_{z}^{m}]$ with the size of $19\times3$. An action can be represented as a set of feature vector of all image as it is shown in Eq. \ref{e4}.
\begin{equation}
\label{e4}
    F = [f_{1},f_{2}, ......f_{M}]
\end{equation}

It should be noted that due the the different heights of human bodies, the numerical value of the coordinate of skeleton will ranges. The taller a person is, the longer the skeletal segment will be. This difference may cause misidentifying of the same action but made by different people. Considering the skeletal segment size, a normalization function is used. As shown in Fig. \ref{fig:meth}, the point 1 is the hip center, and the point 2 is the middle of spine. The final action space feature vector is shown by the Eq. 7.

\begin{align}
    \label{e5}
&\bar{f} = f*Scale  \\ 
&Scale = distance(P_{1},P_{2})/(length\_of\_spine/2) \\
&\bar{F} = [\bar{f_{1}},\bar{f_{2}}, ......\bar{f_{M}}]
\end{align}

\subsection{Model Creating}
\begin{figure}[ht]
    \centering
    \subfigure[Sample RNNs]{
    \label{Fig.sub.1}
    \includegraphics[width=0.2\textwidth]{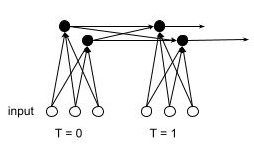}}
    \subfigure[IndRNNs]{
    \label{Fig.sub.2}
    \includegraphics[width=0.2\textwidth]{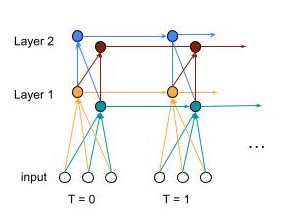}}
    \caption{Architecture of Simple RNN and IndRNN unfloded in time.}
    \label{Fig.SInd}
\end{figure}

%
RNN is a class of deep neural networks which take not only the current input but also the previous hidden state as the true input. Depending upon the number of time steps, RNN can efficiently retain information about the past. Therefore, RNN is more suitable for time-series analysis. However, RNN and its variants Long short-term memory (LSTM) network are easily prone to long-range dependency problem, such as gradient explode and vanishes during back-propagation operations. To overcome such issues, IndRNN is proposed \cite{li2018independently} as an enhanced version of simple RNNs. In IndRNNs the neurons in the same layer are independent of each other while connected across layers. Figure \ref{Fig.SInd} shows the architecture differences between simple RNNs and IndRNNs. In a particular hidden layer of IndRNN as shown in Fig. \ref{Fig.sub.2}, each neuron only receive its own past context information instead of having full connectivity among all neurons in the same layer as the simple RNN does as shown in Fig. \ref{Fig.sub.1}.  

\begin{figure}[ht]
  \centering
  \includegraphics[width=0.65\linewidth]{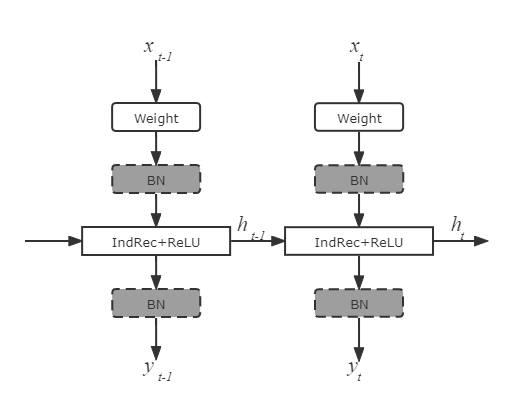}
  \caption{Basic architecture of IndRNN.}
  \label{fig:ind}
\end{figure} 

In this work, an IndRNN architecture \cite{li2018independently} is adopted to handle the HAR task. The basic architecture is represent in Fig. \ref{fig:ind}. The \emph{IndRec+ReLU} block represents the input and recurrent process carried out at each time step with ReLU activation function. \emph{BN} donates the batch normalization preformed before and after the activation function \cite{li2018independently}. The \emph{Hadamard} product is used to process the recurrent inputs in the hidden layer of the processed architecture. For each time step $t$, the $n_{th}$ hidden state $h_{n,t}$ can be updated with Eq. \ref{hadamard},
\begin{equation}
\label{hadamard}
    h_{n,t} = \sigma (W_{n}x_{t}+u_{n} \odot h_{n,t-1}+b_{n})
\end{equation}
where $W_{n}$ is a vector represents the input weight, $u_{n}$ is the recurrent weight, $\sigma$ is the ReLU activation function, $\odot$ denotes the hadamard product, and $b_{n}$ is the bias. After being trained, the IndRNN-based HAR model is able to handle the extracted skeleton features.
The 4-layer IndRNN structure has achieved comparable experimental results \cite{li2018independently}.
Particularly, the IndRNN is set to be a 4-layer simple structure to meet the requirement of smart-home environment. 

\subsection{Real-Time Testing and Sliding Windows}
\begin{figure}[ht]
  \centering
  \includegraphics[width=0.75\linewidth]{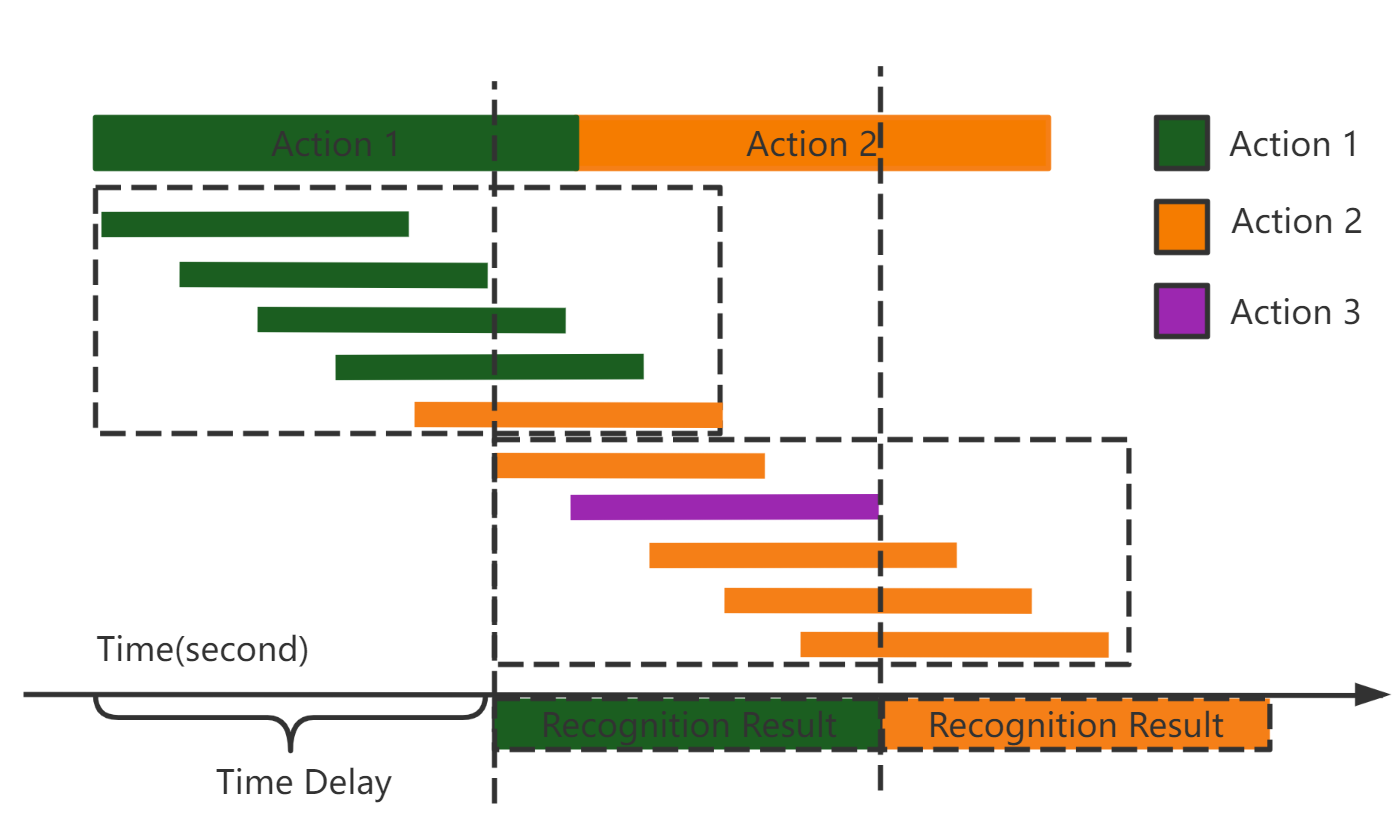}
  \caption{Sliding window method for action recognition.}
  \label{fig:sliding}
\end{figure}

Most work on continuous HAR assumes that pre-segmented video clips are used in recognition part of the task. However, the information of the start and ending time of an observed action are import in order to provide a credible recognition result for untrimmed action streams. In this paper, we apply a sliding window method to divide the input skeleton video into short overlapped skeleton segments as shown in Fig. \ref{fig:sliding}. Specifically, the input skeleton video streams are sampled every five frames to construct a sequence of frames for processing by a sliding window of 20 frames as time elapses. 

After sliding window video trim, the short sequence is fed into the action recognition system to estimate the result. But the sliding window method greatly increases the result number, which consequently will reduce the final accuracy and cause confusion at the junction of two actions. Besides, in real-world applications, such a high frequency result report is unnecessary. Therefore, Non-maximum suppression (NMS) method is adopted to increase the robustness of the system. In every five sliding windows, the sliding results are utilized to find the real action result and provide the start and stop time. The result is annotated at the Starting time of $6$-th sliding window.  

\section{Experimental Results}
\label{result}

\subsection{Experimental Setups}

Experimental study has been conducted to evaluate the proposed REMS scheme,and the results are analyzed in terms of the training and testing accuracy using IndRNN architecture and mean Average Precision of the instance-level action recognition. The model is trained on an Intel i7-8700k \@ 3.7GHz and two NVIDIA GeForce GTX1080Ti graphic cards under Windows 10. The experiments are conducted under python 3.8 that utilizes Pytorch backend and NVIDIA CUDA 11.6 library for parallel computing.  

\subsection{Dataset: NTU\_RBG+D}

In this study, the training process is conducted on the NTU\_RGB+D dataset, which has 60 action classes performed by 40 subjects from 80 views \cite{shahroudy2016ntu}. Each action is captured by three Microsoft Kinect V2 places on different positions, the available data for each frame includes RGB image, depth map sequence, 3D skeleton data, and infrared (IR) data. The skeleton data contains 3D coordination of 25 body joints for each frame. In order to satisfy the feature extraction requirement, the last five joints are ignored, which are 21 (top of spine), 22 (tip of left hand), 23 (left thumb), 24 (tip of right hand), and 25 (right thumb). It contains 56880 video samples from 60 types of different classes including 2 segmentation methods: Cross-Subject 
40 subjects are splited in to training and testing sets based on the subject id, each group consists of 20 subjects. 40320 and 16560 samples for training and testing, respectively) and Cross-View (All samples from camera 1 are picked for testing and other samples of camera 2 and 3 for training. 37920 and 18960 samples for training and testing, respectively).

Specifically, there are nine types of medical conditions as shown in Table \ref{table:types}. Also, there are 11 types of actions that are interaction among multiple people. Therefore, 48419 video sample of 49 types of action in total are used for training. 20 frames were sampled from each sequence as one input \cite{li2018independently}. After sampling, the skeleton data used to represent one action is only 10KB, while the RGB image and depth image representing the same action are 1.88MB and 878KB respectively. The low memory consumption makes skeleton image more suitable for action recognition conducted by resource constrained edge devices in smart-home environments. 

\begin{table}[]
\caption{Class of actions related to medical conditions}
\begin{tabular}{|l|l|l|l|l|l|}
\hline
Number & Action       & Total & Number & Action         & Total \\ \hline
41     & Sneeze/Cough & 948   & 46     & Back Pain      & 948   \\ \hline
42     & Staggering   & 947   & 47     & Neck Pain      & 948   \\ \hline
43     & Falling Down & 946   & 48     & Nause/Vomiting & 946   \\ \hline
44     & Headache     & 946   & 49     & Fan Self       & 946   \\ \hline
45     & Chest Pian   & 947   &        &                &       \\ \hline
\end{tabular}
\label{table:types}
\end{table}

\begin{table}[]
\centering
\caption{Results on NTU RGB+D dataset}
\begin{tabular}{|l|l|l|l|}
\hline
Method                                      & Year & Cross-Subject & Cross-View      \\ \hline
2-Layer RN\cite{shahroudy2016ntu}                  & 2016 & 56.3    & 64.1    \\ \hline
1-Layer LSTM\cite{shahroudy2016ntu}                & 2016 & 59.1    & 66.8    \\ \hline
2-Layer LSTM\cite{shahroudy2016ntu}                & 2016 & 60.7    & 67.3    \\ \hline
ST-GCN\cite{yan2018spatial}                       & 2018 & 81.57   & 88.84   \\ \hline
4-Layer IndRNN \cite{tsai2020deep}                & 2020 & 77.23   & 88.3    \\ \hline
ST-CNN\cite{wang2021skeleton}                     & 2021 & 80.29   & 85.77 \\ \hline
\textbf{REMS}                                     & 2022 & 79.87    & 89.76 \\ \hline
\end{tabular}
\label{table:train}
\end{table}

Table \ref{table:train} presents the comparisons of REMS with other existing methods. In \cite{tsai2020deep}, the joint coordinates of two subject skeletons are set as the input. If only one person is presented, the other is set as zero. With a different feature extraction method, our REMS system is more focusing on single skeleton analysis in purpose to accommodate the situation of elders who live alone. The accuracy of REMS is slightly lower than the ST-GCN based models \cite{yan2018spatial}, in which the input is spatially processed through Graph convolution. However, in the context of smart home monitoring at the network edge, the available computational resources are limited. Consequently, a lightweight action recognition model with less complexity especially in feature extraction step is more affordable and the trade-off is acceptable. 

\subsection{Continuous Action Recognition Results}
Table \ref{table:TSU} reports the Average Precision (AP) of REMS on activity classification. Beside NTU\_RGB+D, the Toyota Smarthome Untrimmed (TSU) dataset is newly published and contains 536 video streams with an average of 21 minutes, which annotated with 51 activities. The subjects are seniors in the age range from 60-80 years old. Therefore, this dataset is a qualified to represent the environment of elderly living alone. The sliding window overlapping is chosen to be 75\% while the window length is 20 as the training data size as suggested in \cite{shou2016temporal}. The use of two different datasets is also purposed to be closer to the real-world situations. In real life, human actions are often unpredictable and actions that are not included in the training set are very likely encountered. There are total eight types of actions that occur in both dataset at the same time.  

\begin{table}[]
\centering
\caption{Average Precision (AP) of REMS on activity classification using TSU and NTU\_RGB+D dataset}
\begin{tabular}{|l|l|l|l|l|}
\hline
Action Type & Drinking & Eat snack & Sitting Down & Stand up  \\ \hline
NTU\_RGB+D  & 0.8660   & 0.6608    & 0.8521       & 0.6119    \\ \hline
TSU         & 0.1984   & 0.2637    & 0.3502       & 0.4025    \\ \hline
Action Type & Reading  & Writing   & Use glasses  & Phonecall \\ \hline
NTU\_RGB+D  & 0.5401   & 0.5728    & 0.7706       & 0.9666    \\ \hline
TSU         & 0.2376   & 0.2901    & 0.2354       & 0.4959    \\ \hline
\end{tabular}
\label{table:TSU}
\end{table}
The start-of-art mAP of the TSU is 0.327 \cite{dai2021pdan} while the mAP over selected action types is 0.308, which shows that our REMS scheme achieved a comparable accuracy. However, it is worthy to mention that only eight types of actions are contained in TSU and NTU\_RGB+D dataset both, and REMS is trained using NTU\_RGB+D dataset only. This experiment is conducted to evaluate the performance of REMS in case it is deployed to handle actions that are not in the training data set. This study leads to some interesting observations. Although there are identically labeled actions in both dataset, the results vary widely due to multiple factors, e.g. different camera settings. Actions that do not appear in the training set are very mislabeled, which lead to a question on how to avoid that situation in real-life applications. Finally, the healthcare-related actions are not given in this dataset and there is very few untrimmed dataset related to healthcare. Hence, introducing artificially generated action data using tools such as GANs might be feasible in the next step. 


\section{Conclusions}
\label{Conclusion}

In this work, we propose REMS, a real-time elderly monitoring for senior safety by applying lightweight human action recognition system. Leveraging a plain IndRNN structure as the main action recognition core, the REMS system is efficient and is ready to be transplanted to edge devices that are affordable in smart-home environments. There are time step delays after the implementation of using NMS. Considering that the Kinect sensor works around 30 frames per second (FPS), the time delay is around five seconds. This delay does not significantly impact the healthcare monitoring because the time that emergency personnel arrives after receiving the alarm is measured in minutes. Certainly shorter delay time will be more approving, which is also included in our future work.

Privacy and security are among the top concerns that need to be considered in tracking and monitoring process especially in the application scenarios like healthcare monitoring \cite{dai2020toyota}. In our REMS system, only the skeleton information is used. REMS does not involve any privacy sensitive information, such as the living environments, the color and style of clothing, items held in the person's hand, etc.. Regarding the data that is closely related to personal information such as height, the personal tag can be weakened after the normalization in feature extraction step. Therefore, even if the data is leaked into cyberspace, it will not bring any impact on personal privacy.

\bibliographystyle{IEEEtranS}
\bibliography{HAR}

\end{document}